\documentclass[conference]{IEEEtran}

\IEEEoverridecommandlockouts

\usepackage{amsmath,amssymb,amsfonts}
\usepackage{algorithm}
\usepackage{algpseudocode} 
\usepackage{graphicx}
\usepackage{textcomp}
\usepackage{xcolor}
\usepackage{booktabs}
\usepackage{multirow}
\usepackage{array}
\usepackage{listings}
\usepackage{url}
\usepackage{subcaption}

\usepackage[numbers]{natbib}
\newcommand{\fname}{ARASH}

\usepackage{tikz}
\usetikzlibrary{shapes.geometric}

\newcommand{\circlednumber}[1]{\tikz[baseline]\node[circle,draw=black,fill=black,text=white,inner sep=1pt,minimum size=1pt,anchor=base] {#1};%
}

\lstdefinestyle{onecolcode}{
  basicstyle=\ttfamily\footnotesize,
  columns=fullflexible,
  breaklines=true,
  breakatwhitespace=true,
  frame=single,
  showstringspaces=false,
  keepspaces=true
}

\begin{document}

\title{\fname: Adaptive Retrieval And Shot Selection for Tabular Prediction }

\author{
    \IEEEauthorblockN{Samirasadat Jamalidinan, Yue Xu, and Kazem Cheshmi}
    \IEEEauthorblockA{\textit{Department of Electrical and Computer Engineering} \\
    \textit{McMaster University}\\
    Hamilton, ON, Canada \\
    Email: \{jamalids, xuy280, cheshmi\}@mcmaster.ca}
}

\maketitle

\begin{abstract}

Tabular prediction is a critical task across numerous applications. The recent success of large language models has sparked various approaches for adapting them to the tabular domain. A prevalent strategy involves training or fine-tuning specialized Tabular Foundation Models (TFMs) such as TabPFN. However, TFMs require substantial computational resources, and frequent model retraining is often impractical. In-context learning (ICL), specifically, few-shot prompting, offers a resource-efficient alternative to enhance performance. Yet, identifying the most relevant rows to serve as shots remains a challenge for tabular data. This paper introduces ARASH (Adaptive, query-specific Retrieval And Shot selection), a method that improves TFM efficiency by selecting optimal shots based on local neighborhood analysis within the training set. Our results demonstrate that ARASH reduces the prompt length and memory usage of TabPFN by 1261.5$\times$ and 2.56$\times$, respectively, while providing comparable accuracy.

% enhances the performance of major TFMs, including TabPFN, across various tabular prediction tasks.

\end{abstract}

\section{Introduction}

%% tabular task
Tabular prediction is a critical task involving the estimation of missing information in structured datasets. This task is foundational to various domains, including finance and healthcare, where individual records with missing features or labels are provided as input for inference. While deep learning has revolutionized many fields, neural models~\cite{wang2021dcnv2} often struggle with the inherent heterogeneity and mixed modalities of tabular data, creating significant architectural challenges.
The traditional approach to this task relies on classical tree-based ensemble methods, such as XGBoost~\cite{chen2016xgboost}, which efficiently predict tabular data.

% LLMs and ICL is good but it is sensitive to data
% The success of large language models (LLMs) across diverse domains has introduced a flexible alternative through in-context learning (ICL) \citep{brown2020language}. LLMs can perform supervised prediction in a few-shot manner by conditioning on interleaved input-label demonstrations. In this paradigm, model parameters remain fixed, and task adaptation occurs entirely through the content and structure of the prompt~\cite{brown2020language}. 
% %
% However, applying these techniques to the tabular domain remains difficult, as different tables define structures and features in distinct ways and often lack a straightforward transformation into sequential text.
% Empirical studies in non-tabular domains indicate that ICL performance is highly sensitive to the specific demonstrations selected, the method of textualization, and the number of shots allocated \citep{liu2022makes}. Retrieval-based prompting has consequently emerged as a practical mechanism to automate demonstration selection by identifying labeled instances similar to the query within an embedding space \citep{shi2022knn,xu2023k}. 

The success of language models (LMs) across diverse domains has introduced a flexible alternative through in-context learning (ICL) \citep{brown2020language}. Large LMs (LLMs) can perform supervised prediction in a few-shot manner by conditioning on interleaved input-label demonstrations. In this paradigm, model parameters remain fixed, and task adaptation occurs entirely through the content and structure of the prompt~\cite{brown2020language}. 
Empirical studies in non-tabular domains indicate that ICL performance is highly sensitive to the selected demonstrations, the method of textualization, and the number of shots allocated \citep{liu2022makes}. Retrieval-based prompting has consequently emerged to automate demonstration selection by identifying labeled instances similar to the query within an embedding space \citep{shi2022knn,xu2023k}.

Tabular foundation models (TFMs), such as TabPFN~\cite{hollmann2022tabpfn}, TabDPT~\cite{ma2024tabdpt}, and TabLLM~\cite{hegselmann2023tabllm}, have demonstrated strong performance by enabling ICL tailored to tabular data; these models are typically either pre-trained on large-scale tabular collections~\cite{hollmann2022tabpfn,ma2024tabdpt} or fine-tuned using downstream data~\cite{hegselmann2023tabllm}, and rely on carefully constructed prompts that aggregate relevant contextual information for each query. A common strategy is to use the full training set as context, which provides comprehensive coverage but incurs substantial computational cost, with models such as TabPFN exhibiting quadratic memory scaling with respect to the number of shots~\cite{hollmann2022tabpfn}. To mitigate this, prior work has explored few-shot selection using either fixed numbers of examples~\cite{narayan2022can} or locality-aware retrieval methods such as k-Nearest Neighbors (kNN)~\cite{xu2023k}, aiming to balance efficiency and accuracy. 

However, while locality-based approaches exploit the observation that predictive patterns are often stronger within local regions of the feature space~\cite{basu2023statistical,thomas2024retrievalfinetuningincontext}, they introduce new challenges: the appropriate value of $k$ is unclear and dataset-dependent, and 
local neighborhoods may suffer from label impurity, especially when class imbalance makes minority-class examples sparse~\cite{he2009learning}, resulting in noisy or unrepresentative context. Consequently, despite leveraging locality, ICL performance often remains below that of full-data training, highlighting the need for query-specific strategies that jointly account for locality and label purity.

%% ARASH
This paper introduces Adaptive, query-specific Retrieval And Shot selection (ARASH), a novel ICL technique that selects and retrieves shots based on a specific query for tabular prediction. ARASH finds local regions with pure labels to select the minimum number of representative shots. 
ARASH first identifies local regions relative to the query within the training set using a clustering technique. It then employs a difficulty-aware shot selection process to determine the optimal number of shots based on locality and purity of the query's neighborhood. Finally, shot retrieval is performed from the selected clusters based on locality and purity.
The retrieved shots are subsequently passed to a pre-trained TFM. ARASH, when combined with TabPFN, provides accuracy competitive with TabPFN while reducing the memory usage by 2.56$\times$.

%Figure~\ref{fig:prompt-length-accuracy} demonstrates that ARASH achieves comparable average accuracy to TabPFN while incurring significantly lower API costs \cite{tabpfn_client} due to a reduced token length. 
\section{Motivation}
\label{sec:motiv}

\begin{figure}[!t]
    \centering
    \includegraphics[width=\linewidth]{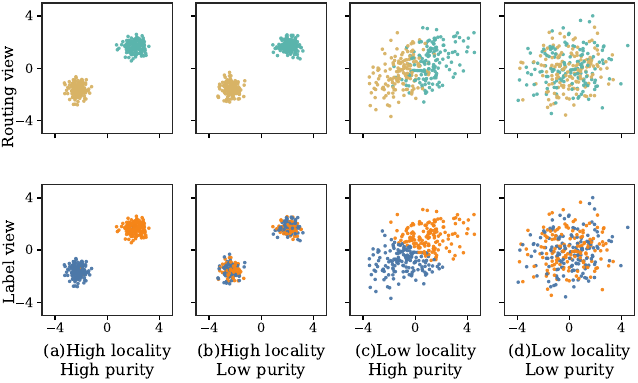}

    \caption{
    Motivation example for \fname{}. Four synthetic regions are shown. Each point represents one demonstration with two features, shown on the horizontal and vertical axes. The colors in the top row denote routed local regions. The bottom row shows the same points, where colors denote class labels. %The comparison illustrates that locality and purity are distinct: a region may be geometrically local while label-mixed, or label-consistent while weakly local.
    }
    \label{fig:arash_motivation}

    \vspace{1.0em}

    \captionsetup{type=table}
    \resizebox{\columnwidth}{!}{%
        \setlength{\tabcolsep}{3pt}
        \begin{tabular}{lcccccccc}
            \toprule
            & \multicolumn{2}{c}{\textbf{Figure 1a}} 
            & \multicolumn{2}{c}{\textbf{Figure 1b}} 
            & \multicolumn{2}{c}{\textbf{Figure 1c}} 
            & \multicolumn{2}{c}{\textbf{Figure 1d}} \\
            \cmidrule(lr){2-3} 
            \cmidrule(lr){4-5} 
            \cmidrule(lr){6-7} 
            \cmidrule(lr){8-9}
            \textbf{Method} 
           & Acc. & Demos.
            & Acc. & Demos. 
            & Acc. & Demos.
            & Acc. & Demos. \\
            \midrule
            Full context & 0.9625 & 3280 & 0.9500 & 3280 & 0.9250 & 3280 & 0.5125 & 3280 \\
            kNN          & 0.9625 & 32   & 0.9300 & 32   & 0.9260 & 32   & 0.4400 & 32   \\
            \fname{}     & 0.9625 & 6    & 0.9500 & 19   & 0.9300 & 13   & 0.5050 & 58   \\
            \bottomrule
        \end{tabular}
    }

    \caption{
    Accuracy (Acc) and number of demonstrations (Demos) for the methods across regions shown in Figure~\ref{fig:arash_motivation}.
    }
    \label{tab:arash_motivation_regions}
\end{figure}

To explain locality and purity, and to show that locality alone is insufficient for reliable local prediction in TFMs, we use four synthetic data patterns (Figure~\ref{fig:arash_motivation}). Each dataset has two features, two labels, and 3280 demonstrations, with feature space (top) and label distribution (bottom) visualizations. Classes are color-coded. We evaluate three shot-selection strategies: (i) Full-context, which uses all $N=3280$ demonstrations; (ii) kNN (locality-aware), which selects fixed $k=32$ nearest demonstrations; and (iii) ARASH (locality- and purity-aware), which adaptively selects both the $k$ value and samples based on region characteristics. The outputs of all methods, combined with a query, are passed as a prompt to TabPFN. We only report accuracy and the number of selected demonstrations. To ensure reproducibility, 80 queries are generated for each data region, and the average accuracy is reported. The query is ``read the demonstration and output exactly one label from the true/false list''.

Full-context TabPFN uses the entire training set without adapting to local query structure. This increases computational and token costs, scaling quadratically and linearly, respectively. kNN reduces this cost by 103 times by selecting local demonstrations from compact geometric regions. However, locality alone is insufficient because such regions may have different purity levels. For example, Figures~\ref{fig:arash_motivation}a and ~\ref{fig:arash_motivation}b both show local regions, but the former is pure while the latter contains mixed labels. In high-locality, low-purity settings, nearby samples provide unreliable label evidence, reducing the accuracy by about 2--3\% compared to when high locality and purity or full-context. %But kNN reduces computation by up to five times compared to full training. 

ARASH addresses this limitation by selecting shots based on both locality and purity. Unlike kNN, which implicitly assumes high purity and thus degrades in low-purity cases, ARASH adapts to such conditions and improves accuracy (by approximately 2--6\% in low-purity cases, Figure~\ref{fig:arash_motivation}b and \ref{fig:arash_motivation}d). Furthermore, in highly pure regions, ARASH enables direct label estimation without invoking the tabular foundation model, as shown in Figure~\ref{fig:arash_motivation}a. Thus, purity not only measures the reliability of local evidence but also identifies cases where expensive model inference can be avoided.

\section{ARASH}
\label{sec:arash}
The goal of ARASH is to adaptively select a compact, query-specific set of informative demonstrations for tabular prediction with TFMs and LLMs. Unlike fixed local retrieval, ARASH jointly considers feature-space locality and label purity to determine when local demonstrations are reliable. This section first formalizes the problem setting and then presents the three stages of the ARASH algorithm.
\subsection{Problem Definition and Overview}
Few-shot prompting enables TFMs and large language models to predict a query label by conditioning on a set of task-specific demonstrations. Given a classification task $\mathcal{T}$ and an input query $x_q$, the model predicts a class label $y_q$ conditioned on a shot set $Shots_{x_q}=\{s_1,s_2,\ldots,s_n\}$:
\begin{equation}
    y_q = \mathrm{TFM}(x_q \mid Shots_{x_q}),
    \label{eq:obj}
\end{equation}
where each demonstration $s_i=(x_i,y_i)$ consists of a tabular instance and its ground-truth label.
Let $\mathcal{D}_{tr}=\{(x_i,y_i)\}_{i=1}^{N}$ denote the training set with feature set $\mathcal{F}=\{f_j\}_{j=1}^{D}$. Each demonstration $x_i=(r_{i1},\ldots,r_{iD})$ contains numerical, categorical, or textual feature values. The objective is to construct, for each query $x_q$, a compact demonstration in $Shots_{x_q}$ that maximizes the accuracy of the prediction while reducing the number of demonstrations passed to the TFM. 

As shown in Algorithm~\ref{alg:ss}, ARASH computes $Shots_{x_q}$ for the input query $x_q$ and $\mathcal{D}_{tr}$ in three stages. ARASH also takes bounds for the number of shots, $k_{min}$ and $k_{max}$, as well as thresholds for locality and purity, i.e., $\tau_{\mathrm{loc}}$ and $\tau_{\mathrm{pur}}$.  ARASH first profiles the training data and partitions it into a set of clusters $\mathcal{C}$, where each cluster represents a local region of the feature space. Second, it assigns a shot budget $k_c$ to each cluster $c\in\mathcal{C}$ using a difficulty score based on local uncertainty and label impurity; clusters with greater estimated difficulty receive larger shot budgets. Third, given a test query $x_q$, ARASH routes the query to a cluster $c_q$ and selects demonstrations according to two complementary reliability diagnostics: dataset-level locality and cluster-level purity. 
\algnewcommand{\Inputs}{\item[\textbf{Inputs:}]}
\algnewcommand{\Outputs}{\item[\textbf{Outputs:}]}

\begin{algorithm}[h]
\caption{ARASH Algorithm}
\label{alg:ss}
\begin{small}
\begin{algorithmic}[1]

\Inputs 
Labeled training set $(X_{\mathrm{tr}},y_{\mathrm{tr}})$ with $N$ rows;\\
Query $x_q$; shot bounds $k_{\min}, k_{\max}$;\\
Locality threshold $\tau_{\mathrm{loc}}$; purity threshold $\tau_{\mathrm{pur}}$

\Outputs 
Shot set $Shots_{x_q}$ %and inference mode $m_q$

\State \textcolor{olive!35}{\textbf{Step I:} Locality-aware clustering }
\State $(Z, \mathrm{L}_{\mathcal{D}}, DP) \gets \mathrm{DataProfiling}(X_{\mathrm{tr}},y_{\mathrm{tr}})$ \label{lin:step1b}
%\State $Z \gets DP.\mathrm{features}$
%\State $L_D \gets DP.\mathrm{locality}$
%\State $r \gets DP.\mathrm{retrieval\_method}$
\State $clustering\_method \gets \mathrm{auto\_selection}(Z, \mathrm{L}_{\mathcal{D}}, DP)$ \label{lin:selection}
\State $\mathcal{C} \gets \mathrm{clustering}(Z, y_{\mathrm{tr}}, clustering\_method)$ \label{lin:step1e}

\State \textcolor{olive!35}{\textbf{Step II:} Per-cluster profiling and shot assignment}
\For{$c \in \mathcal{C}$} \label{lin:step2b}
    \State $\mathcal{Y}_c \gets \{y_i \mid x_i \in c\}$
    \State $H_c \gets \mathrm{NormEntropy}(\mathcal{Y}_c)$
    \State $P_c \gets \mathrm{Purity}(\mathcal{Y}_c)$
    \State $d_c \gets \frac{1}{2}\bigl(H_c + (1 - P_c)\bigr)$
\EndFor\label{lin:percluster}

\State $d_{\min} \gets \min\limits_{c \in \mathcal{C}} d_c$
\State $d_{\max} \gets \max\limits_{c \in \mathcal{C}} d_c$

\For{$c \in \mathcal{C}$}\label{lin:normb}
    \If{$d_{\max} > d_{\min}$}
        \State $\tilde{d}_c \gets \frac{d_c - d_{\min}}{d_{\max} - d_{\min}}$
    \Else
        \State $\tilde{d}_c \gets 0.5$
    \EndIf

    \State $k_c \gets 
    \mathrm{round}\!\left(k_{\min} + \tilde{d}_c (k_{\max} - k_{\min})\right)$\label{lin:bnd1}

    \State $k_c \gets 
    \min\!\Bigl(|c|,\,
    \max\!\bigl(k_c,\ |\mathrm{unique}(\mathcal{Y}_c)|\bigr)\Bigr)$ \label{lin:bnd2}
\EndFor \label{lin:step2e}

\State \textcolor{olive!35}{\textbf{Step III:} Query routing and retrieval}
\State $c_q \gets \mathrm{assign}(x_q, \mathcal{C})$\label{lin:step3b}

\If{$\mathrm{L}_{\mathcal{D}} \geq \tau_{\mathrm{loc}}$ \textbf{and} $P_{c_q} \geq \tau_{\mathrm{pur}}$}\Comment{\textcolor{olive!35}{local \& pure}}\label{lin:ifbeg}

    \State $Shots_{x_q} \gets 
    \mathrm{ClusterShotSelection}(x_q, c_q, k_{c_q})$

\ElsIf{$\mathrm{L}_{\mathcal{D}} \geq \tau_{\mathrm{loc}}$ \textbf{and} $P_{c_q} < \tau_{\mathrm{pur}}$}\Comment{\textcolor{olive!35}{local, not pure}}

    \State $Shots_{x_q} \gets 
    \mathrm{DiverseClusterShotSelection}(x_q, c_q, k_{c_q})$

\ElsIf{$\mathrm{L}_{\mathcal{D}} < \tau_{\mathrm{loc}}$ \textbf{and} $P_{c_q} \geq \tau_{\mathrm{pur}}$}\Comment{\textcolor{olive!35}{not local, pure}}

    \State $Shots_{x_q} \gets 
    \mathrm{HybridShotSelection}(x_q, c_q, k_{c_q}, k_{\max})$

\Else \Comment{\textcolor{olive!35}{not local, not pure}}

    \State $Shots_{x_q} \gets 
    \mathrm{GlobalShotSelection}(x_q, k_{\max})$

\EndIf \label{lin:step3e}

% \State \textcolor{blue}{\textbf{Post-processing}}
% \If{$\left|\{y_i \mid (x_i,y_i) \in Shots_{x_q}\}\right| = 1$}
%     \State $m_q \gets \textsc{Direct}$
% \Else
%     \State $m_q \gets \textsc{TFM}$
% \EndIf

\State \Return $Shots_{x_q}$%$, m_q)$

\end{algorithmic}
\end{small}
\end{algorithm}

\subsection{Step I. Locality-aware Clustering}

The primary objective of the first step is to partition the training dataset $X_{\mathrm{tr}}$ into meaningful local regions to be used in Step II. Recognizing that tabular datasets exhibit heterogeneous geometric and statistical characteristics, ARASH avoids a fixed clustering strategy. Instead, it dynamically fits a clustering model based on a selected normalized feature space to create the clustering set $\mathcal{C}=\{c_1,\ldots,c_t\}$ where each cluster is a local region.  Lines~\ref{lin:step1b}--\ref{lin:step1e} in Algorithm~\ref{alg:ss} show the step I of ARASH.

The $DataProfiling$ function in line~\ref{lin:step1b} in Algorithm~\ref{alg:ss} is applied once to the training set before clustering. The profiling computes a compact summary of structural properties, including dataset size $N$, feature dimensionality $D$, a normalized feature-space representation $Z$, anisotropy measured via PCA eigenvalue ratios, the Hopkins statistic, and some statistics of the above items collectively shown with $DP$.
We use the Hopkins statistic as the dataset-level locality score $\mathrm{L}_{D}$ because it quantifies whether the normalized feature space exhibits meaningful neighborhood structure. Let $u_i$ denote the nearest-neighbor distance from a uniformly sampled point to the training set, and let $w_i$ denote the nearest-neighbor distance from a sampled training point to its nearest neighboring training point. We compute the locality score as:
\begin{equation}
\mathrm{L}_{\mathcal{D}}=H(X_{\mathrm{tr}})=
\frac{\sum_{i=1}^{m}u_i^D}
{\sum_{i=1}^{m}u_i^D+\sum_{i=1}^{m}w_i^D}.
\label{eq:hopkins}
\end{equation}
where $D$ is the feature dimensionality and $m$ is the number of sampled points.
Values near $0.5$ indicate weak locality, while larger values indicate stronger
clustering tendency and stronger support for neighborhood-based retrieval. This locality score is stored in $DP$ and reused during query routing in Step III; it is not recomputed for each query or each cluster. These profiling signals are then used to apply heuristic rules that balance scalability and modeling
flexibility.

The final choice of the clustering algorithm is governed by an automatic clustering selector that evaluates a candidate pool consisting of \texttt{KMeans}, \texttt{MiniBatchKMeans}, \texttt{GMM}, \texttt{Birch}, \texttt{HDBSCAN}, and \texttt{AgglomerativeClustering}. Line~\ref{lin:selection} in Algorithm~\ref{alg:ss} shows the automatic clustering selector that selects a clustering method to be used in line~\ref{lin:step1e}. 
%Rather than relying on an expensive hyperparameter optimization, 
The selector makes a single dataset-level decision based on meta-features extracted by the $DataProfiling$ function—such as dataset size, dimensionality, anisotropy, and locality—to choose a clustering method before inference. The thresholds for these metrics are determined via hyperparameter optimization.
%
%Clustering is computed once per dataset in a preprocessing step and reused at inference time.  
%
%The resulting cluster assignment function is stored and reused to route future queries. For each cluster, ARASH records its label distribution, normalized entropy, purity, impurity, and size. These statistics are later used to determine the cluster-level shot budget and to select the retrieval regime for a query.
%“dynamic”\Kazem{what is this?} denotes the automatic choice of the clustering method based on dataset properties.

\subsection{Step II. Difficulty-aware Shot Assignment}
\label{sec:difficulty_shot_assignment}

\fname{} determines the shot budget for each local region rather than using a fixed number of shots for all queries. Here, the shot budget denotes the
number of in-context demonstrations selected from the training data and included in the query-specific shot set \(Shots_{x_q}\). The goal is to allocate a larger context to clusters whose labels are harder to resolve, while assigning fewer shots to clusters that already provide a reliable local signal. This step is implemented by a difficulty-aware controller, shown in lines~\ref{lin:step2b}--\ref{lin:step2e} of Algorithm~\ref{alg:ss}, which maps each cluster to an adaptive shot budget \(k_c\) within the admissible range \([k_{\min}, k_{\max}]\). Bounds $k_{\min}, k_{\max}$ are determined based on the model token length.

The ARASH algorithm first, in lines~\ref{lin:step2b}--\ref{lin:percluster}, computes a difficulty score \(d_c \in [0,1]\) for each cluster \(c \in \mathcal{C}\). The difficulty score is calculated based on normalized label entropy \(H_c\) and label impurity \(1-P_c\). The purity \(P_c\) is defined as the fraction of shots in cluster \(c\) that belong to its majority class:
\begin{equation}
P_c =
\max_{y\in\mathcal{Y}}
\frac{
\left|\{i : x_i \in c,\; y_i = y\}\right|
}{
\left|\{i : x_i \in c\}\right|
}.
\label{eq:cluster_purity}
\end{equation}
A pure cluster has \(P_c=1\), whereas smaller values indicate stronger label mixing. The difficulty score is then defined as
%\begin{equation}
$d_c = \frac{1}{2}\left(H_c + (1-P_c)\right)$.
%\end{equation}
Here, \(H_c\) captures uncertainty in the cluster label distribution, while \(1-P_c\) measures the absence of a dominant label direction. This choice avoids the ambiguity of class imbalance: a cluster dominated by one class may be highly imbalanced, but it can still be reliable for local prediction. In contrast, impurity reflects the type of label conflict that should increase the number of demonstrations.

After computing \(d_c\) for all clusters, \fname{} normalizes the difficulty values per cluster in lines~\ref{lin:normb}--\ref{lin:step2e}. The normalized score is calculated as:
% \begin{equation}
$\tilde{d}_c =
\frac{d_c-d_{\min}}{d_{\max}-d_{\min}}$
% \end{equation}
where \(d_{\min}\) and \(d_{\max}\) denote the minimum and maximum difficulty scores over \(\mathcal{C}\). If all clusters have the same difficulty, \(\tilde{d}_c\) is set to \(0.5\), giving each cluster a midpoint allocation.
The normalized score is then mapped to an integer shot budget by linear interpolation:
% \begin{equation}
$k_c =
\mathrm{round}\!\left(
k_{\min} + \tilde{d}_c (k_{\max} - k_{\min})
\right)$.
% \label{eq:adaptive_k}
% \end{equation}
Thus, clusters with low difficulty receive budgets closer to \(k_{\min}\), whereas label-mixed clusters receive budgets closer to \(k_{\max}\). Finally, \fname{} applies two constraints: \(k_c\) must be at least the number of distinct labels observed in cluster \(c\), and it cannot exceed the cluster size. These two constraints, shown in lines~\ref{lin:bnd1}--\ref{lin:bnd2}, ensure that the selected context is large enough to represent the local label set while remaining feasible for the model.

\subsection{Step III. Query routing and retrieval}
\label{sec:reliability_retrieval}

The third step of ARASH uses purity and locality metrics to retrieve shots for the input query, as shown in lines~\ref{lin:step3b}--\ref{lin:step3e} in Algorithm~\ref{alg:ss}. 
ARASH first assigns a given query \(x_q\) to a cluster \(c_q \in \mathcal{C}\), line~\ref{lin:step3b}. Then it retrieves $k_c$ demonstrations using one of the four retrieval strategies, which are selected based on dataset-level locality score \(\mathrm{L}_{\mathcal{D}}\) and the purity \(P_{c_q}\) as shown in lines~\ref{lin:ifbeg}--\ref{lin:step3e} in Algorithm~\ref{alg:ss}.
The locality score shows whether the feature space supports meaningful neighborhood-based retrieval. Cluster purity shows whether the assigned local region provides consistent label evidence. %These two signals are used together because locality alone does not guarantee reliable labels, and purity alone does not guarantee that the routing decision is stable.

Based on locality and purity scores, ARASH has four different strategies: 
\noindent
\circlednumber{1} local and pure: the assigned cluster is considered reliable. In this case, \fname{} retrieves \(k_{c_q}\) demonstrations from the assigned cluster using a kNN strategy. This case is the most favorable for local prompt construction because the selected demonstrations are both close to the query and likely to provide a consistent label signal.
\noindent
\circlednumber{2} local and impure: the assigned cluster is geometrically meaningful, but its labels are mixed. In this case, local retrieval is still useful, but selecting only the nearest or redundant examples may give an unstable prompt. Therefore, \fname{} uses a diversity-aware local retrieval policy, i.e., DPP, within the assigned cluster. This allows the selected shots to better represent the ambiguous local label structure.
\noindent
\circlednumber{3} Not local but pure: the assigned cluster has a dominant label, but the feature space does not provide strong support for stable local routing. In this case, using only the assigned cluster may be too restrictive. Therefore, \fname{} uses hybrid retrieval, which combines examples from the assigned cluster (local kNN) with globally retrieved examples (global kNN). This keeps the useful label signal from the cluster while reducing dependence on an uncertain local partition.
\noindent
\circlednumber{4} Not local and impure: the assigned cluster is not reliable for local prompt construction. In this case, \fname{} falls back to global shot selection with budget \(k_{\max}\) using global DPP. %This avoids forcing a local prompt when neither the feature geometry nor the label distribution supports local retrieval.

\section{ARASH Implementation}
\label{sec:imp}
This section details the efficient and portable implementation techniques of ARASH. To minimize overhead, ARASH uses a two-stage approach where preprocessing is performed once for all queries, and the foundation model is called only when essential to reduce inference latency. Additionally, its portable architecture seamlessly supports various LMs.

\subsection{Efficient Processing and Inference} 
\label{sec:cached_state}

ARASH is implemented as a two-stage pipeline that cleanly separates dataset-level preprocessing from query-time inference. In the preprocessing stage, all computations that depend solely on the training data are performed once, including feature normalization, data profiling, clustering, cluster statistics, and allocation of cluster-level shot budgets. The outputs of this stage form reusable artifacts that are leveraged during inference. Specifically, ARASH stores (1) the feature schema and normalization parameters, such as categorical levels and scaling statistics, to ensure that test queries are encoded consistently with the training data; (2) the cluster assignment function along with cluster-level label statistics; and (3) a shot-budget table that maps each cluster $c$ to an integer budget $k_c$. During the inference stage, these cached artifacts are reused to efficiently process each query by determining its cluster, loading relevant demonstrations, and producing a direct prediction or invoking a frozen foundation model. This separation is crucial in tabular prediction tasks, where many queries are typically evaluated on the same dataset, allowing ARASH to amortize the one-time cost of profiling and clustering instead of repeating it for every test instance.

In addition to reducing the number of shots to reduce the computational overhead of inference, ARASH also performs a post-processing to skip calling the model when unnecessary. 
After the shots are retrieved, \fname{} applies a lightweight post-processing step. If all retrieved demonstrations have the same label, the query is assigned to \textsc{Direct} mode, and the unique label is returned without calling the TFM. Otherwise, the query is assigned to \textsc{TFM} mode, and the selected demonstrations are passed to the foundation model. This step keeps \fname{} as a retrieval and prompt-construction method: it directly predicts only when the retrieved evidence is already label-consistent, and otherwise leaves prediction to the TFM.

\subsection{Language-model Serialization}
\label{sec:lm_serialization}

To test whether the retrieval policy transfers beyond table-native
foundational models, we also evaluate ARASH with pretrained language models.
We consider LLaMA~3.2 and Qwen~2.5, which are decoder-only models, and
FLAN-T5, which follows an encoder--decoder architecture. These experiments
require an additional serialization step because language models consume text
rather than native tabular tensors.

Each row is converted into a key--value representation
%\begin{equation}
    $z(x_i) = \{f_1 : r_{i1}, f_2 : r_{i2}, \dots, f_D : r_{iD}\}$
%\end{equation}
where \(f_j\) denotes the \(j\)-th feature and \(r_{ij}\) denotes the
corresponding value for row \(i\). Retrieved demonstrations are serialized as input--label pairs, and concatenated with the serialized query. A short task-specific instruction is then appended to form the final prompt. Thus, for language-model backbones, ARASH controls which demonstrations are included in the prompt, while the serialization layer converts the selected
tabular context into a text format accepted by the model.

\section{Experimental Results}
This section evaluates ARASH and provides a comprehensive comparison with prior work. It examines the integration of ARASH with language models and concludes with a detailed technical breakdown of the ARASH algorithm.

\subsection{Setup}

% \subsubsection*{Datasets and environment} We evaluate ARASH on a combined benchmark consisting of OpenML-CC18 \cite{bischl2017openml} and Combo~\cite{fang2024large} datasets.
% The Combo dataset collection provides a compact yet diverse benchmark for tabular classification, enabling controlled comparisons while keeping computational cost manageable.
% We use OpenML-CC18  and Combo set for overall performance evaluation, but for most analysis and ablation studies, we rely on the Combo datasets to keep the evaluation time bounded. All experiments are conducted on a single consumer-grade NVIDIA GeForce RTX 4090 GPU.
\subsubsection*{Datasets and environment}
We evaluate ARASH on a combined benchmark consisting of OpenML-CC18~\cite{bischl2017openml} and Combo~\cite{fang2024large} datasets.
The Combo dataset collection provides a compact yet diverse benchmark for tabular classification, enabling controlled comparisons while keeping computational cost manageable.
We use OpenML-CC18 and Combo for the overall performance evaluation, while most analysis and ablation studies are conducted on the Combo datasets to keep the evaluation time bounded. For datasets, we use an 80/20 train--test split with random seed 42, and the same split is used consistently across all compared methods.
All experiments are conducted on a NVIDIA GeForce RTX 4090 GPU.
\begin{figure}[!t]
  \centering
  \includegraphics[width=\linewidth]{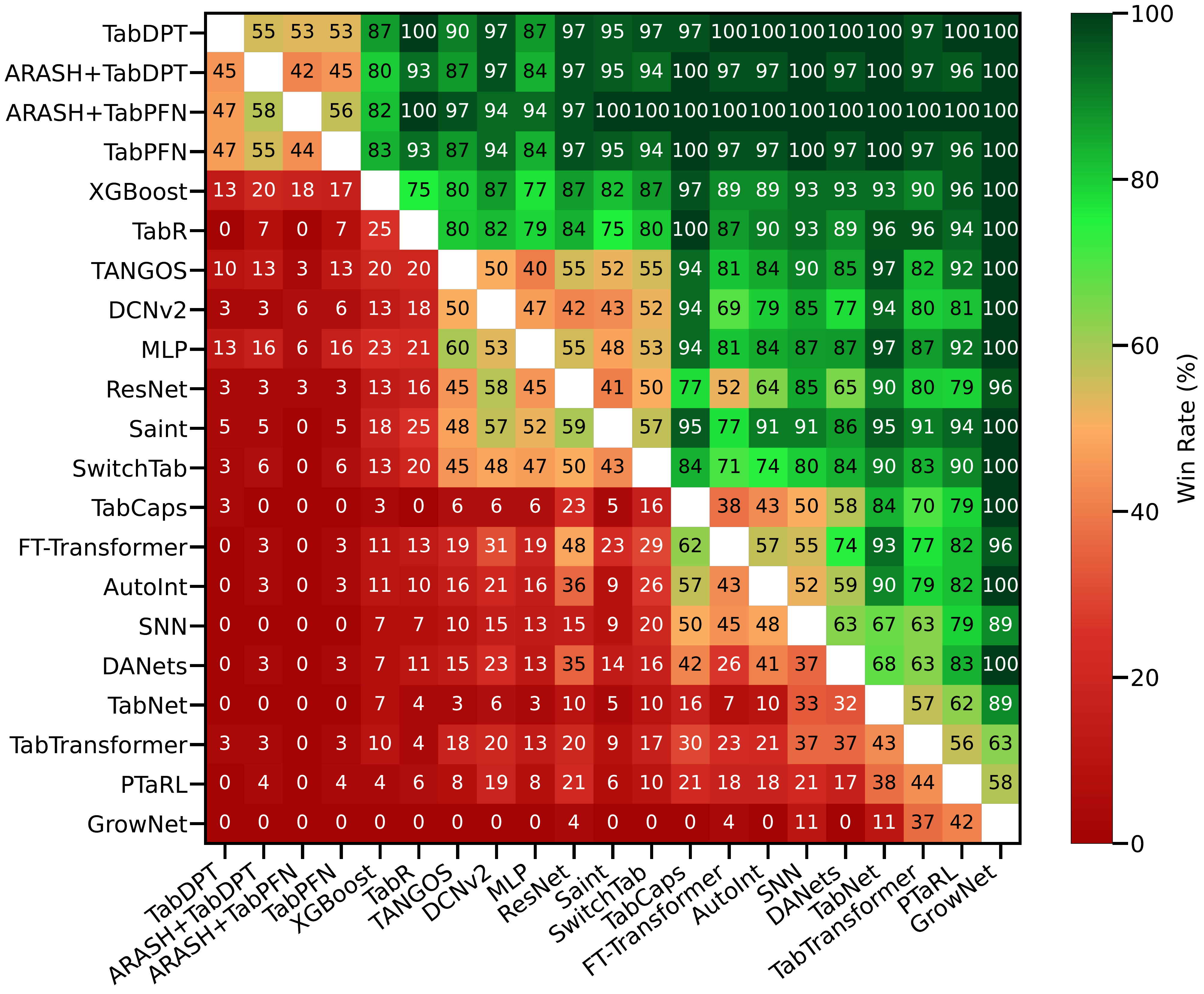}
  %\includegraphics[width=0.4\linewidth]{figs/vram_boxplot_allk_vldb.pdf}
  %\vspace{2pt}
  %\makebox[0.55\linewidth][c]{a) pairwise win-rate }
  %\makebox[0.4\linewidth][c]{b) average accuracy}
  \caption{Pairwise win-rate comparison across baseline models and datasets. % in the TALENT benchmark on the combination of CC18 and Combo datasets. %\Yue{Should we clarify which datasets are used in Fig 2?No} 
  A win is defined as achieving higher classification accuracy on a given dataset.}
  \label{fig:ARASH_win_rate_heatmap}
\end{figure}

% \begin{table}[t]
% \centering
% \small
% \caption{Performance comparison across  on CC18 and Combo datasets.}
% \label{tab:TALENT_avg_acc_F1_table}
% \begin{tabular}{ccc}
% \toprule
% Model & Avg. Accuracy & Avg. MacroF1 \\
% \midrule
% ARASH{+}TabDPT & 0.9016 & 0.800 \\
% TabDPT & 0.9008 & 0.795 \\
% ARASH{+}TabPFN & 0.895 & 0.800 \\
% TabPFN & 0.8948 & 0.799 \\
% XGBoost & 0.879 & 0.785 \\
% TabR & 0.879 & 0.449 \\
% TANGOS & 0.852 & 0.379 \\
% DCNv2 & 0.849 & 0.448 \\
% MLP & 0.849 & 0.379 \\
% Saint & 0.835 & 0.384 \\
% SwitchTab & 0.832 & 0.310 \\
% ResNet & 0.832 & 0.588 \\
% TabCaps & 0.780 & 0.395 \\
% AutoInt & 0.776 & 0.259 \\
% FT-Transformer & 0.771 & 0.216 \\
% DANets & 0.744 & 0.134 \\
% SNN & 0.725 & 0.525 \\
% TabNet & 0.710 & 0.291 \\
% PTaRL & 0.619 & 0.161 \\
% TabTransformer & 0.587 & 0.172 \\
% GrowNet & 0.573 & 0.240 \\
% \bottomrule
% \end{tabular}
% \end{table}

\subsubsection*{Baselines} We compare \fname{} against representative tabular learning baselines from the TALENT benchmark~\cite{liu2025talent}.
TALENT provides a unified evaluation suite for tabular classification and includes standardized implementations and hyperparameter settings for a broad set of deep tabular models. We use the latest TALENT release and report results on the datasets for which the corresponding methods are successfully executed.
TALENT has covered several baseline models, such as 
I) FT-Transformer~\cite{gorishniy2021revisiting}, TabTransformer~\cite{huang2020tabtransformer}, SAINT~\cite{somepalli2021saint}, and AutoInt~\cite{song2019autoint} utilize self-attention mechanisms to explicitly model the contextual relationships between different columns in a table.
II) MLP~\cite{gorishniy2021revisiting}, ResNet~\cite{gorishniy2021revisiting}, and SNN~\cite{klambauer2017snn} from the classical foundation, using basic multi-layer structures and skip connections to process flattened tabular features.
III) DCNv2~\cite{wang2021dcnv2} and DANets~\cite{chen2021danets} are engineered to capture high-order feature overlaps, which are designed to capture higher-order feature interactions.
IV) TabNet~\cite{arik2021tabnet}, GrowNet~\cite{badirli2020grownet}, and TabCaps~\cite{chen2023tabcaps} mimic the logic of decision trees or use specialized ``capsules'' to provide structured, often more interpretable, deep learning pathways.
V) TANGOS~\cite{jeffares2023tangos}, PTaRL~\cite{ye2024ptarl}, and SwitchTab~\cite{wu2024switchtab} apply advanced training techniques like gradient orthogonalization and asymmetric encoding to help standard neural nets generalize better on noisy tables.
VI) TabPFN~\cite{hollmann2022tabpfn} and TabDPT~\cite{ma2024tabdpt} are the TFM that support ICL.
VII) TabR~\cite{gorishniy2024tabr} leverages learned neighbour relationships in an embedding space to improve prediction performance.
\subsubsection*{Evaluation Metrics}
\label{sec:metrics}
 %We evaluate classification performance across datasets with different class distributions. 
 To provide a consistent view of model behavior, we report both average accuracy and Macro-F1. Macro-F1 complements accuracy by reflecting performance across classes with varying frequencies. For ARASH, we define a tuning set as random 10\% of datasets (seed=42) never appearing in training, evaluation, or test and perform a grid search for locality, purity, and auto-clustering thresholds to avoid bias.
%$\tau_{\mathrm{loc}} = 0.80$ and $\tau_{\mathrm{pur}} = 0.5$ to optimize accuracy across the validation combo dataset, and then maintained these parameters across all other datasets.
%
 For DPP, we use the standard quality-diversity decomposition of an $L$-ensemble kernel and apply a fixed-size greedy MAP approximation to obtain the final subset~\cite{kulesza2012determinantal}.

% \subsection{Overall Performance}
% We compare \fname{} against standard tabular baselines and ICL demonstration-selection baselines, focusing on two criteria: predictive accuracy and inference-time VRAM usage.

% \subsubsection{Accuracy} We present a pairwise win-rate comparison of the accuracy between {ARASH+TabDPT}, {ARASH+TabPFN} and all other baseline models in Figure~\ref{fig:ARASH_win_rate_heatmap}. The number in each row represents the win-rate (in percentage) of a specific model against the other model. A "win" is considered for the models with higher prediction accuracy across the OpenML-CC18 and Combo datasets. Table~\ref{tab:TALENT_avg_acc_F1_table} summarizes the average accuracy and Macro-F1 across all baseline models. As illustrated, {ARASH+TabDPT} achieves the highest overall win-rate, followed by TabDPT, {ARASH+TabPFN} and TabPFN.
% % It demonstrates that ARASH can achieve accuracy highly competitive with other Full-context baseline models while requiring only a very limited prompt length, highlighting the effectiveness of the ARASH ICL technique in maintaining performance while reducing computational overhead.
% The results show that \fname{} achieves accuracy comparable to full-context inference while using substantially fewer demonstrations and reducing inference-time memory usage.
\subsection{Tabular baselines}
We present a pairwise win-rate comparison of accuracy between \fname{} with TabDPT and TabPFN and the baseline models in Figure~\ref{fig:ARASH_win_rate_heatmap}. Each entry reports the percentage of datasets on which the model in the row achieves higher prediction accuracy than the model in the column across datasets. The heatmap shows that the TabDPT- and TabPFN-based methods obtain the strongest overall pairwise performance. In particular, \fname{}+TabPFN and TabDPT achieve the highest average pairwise win rates, while \fname{}+TabDPT and TabPFN remain highly competitive. Across the remaining baselines, the \fname{} variants consistently outperform most non-foundation-model methods, indicating that adaptive retrieval preserves the accuracy of tabular foundation models while reducing the number of demonstrations.

In terms of average accuracy, \fname{}+TabDPT obtains the highest accuracy, with an average accuracy of $0.9017$ and Macro-F1 of $0.800$. This is slightly higher than full-context TabDPT, which achieves $0.9009$ accuracy and $0.795$ Macro-F1. Similarly, \fname{}+TabPFN achieves $0.895$ accuracy and $0.800$ Macro-F1, closely matching full-context TabPFN, which obtains $0.8948$ accuracy and $0.799$ Macro-F1. Among the non-foundation-model baselines, XGBoost is the strongest competitor, with $0.879$ accuracy and $0.785$ Macro-F1, followed by TabR with $0.879$ accuracy but a lower Macro-F1 of $0.449$. Other neural tabular baselines, including TANGOS, DCNv2, MLP, SAINT, and SwitchTab, obtain lower average accuracy and substantially lower Macro-F1.

\subsection{ICL Demonstration-Selection Baselines}
\label{sec:icl_baseline_comparison}

\begin{figure}[t]
    \centering

    \begin{subfigure}[t]{0.48\textwidth}
        \centering
        \includegraphics[width=\linewidth]{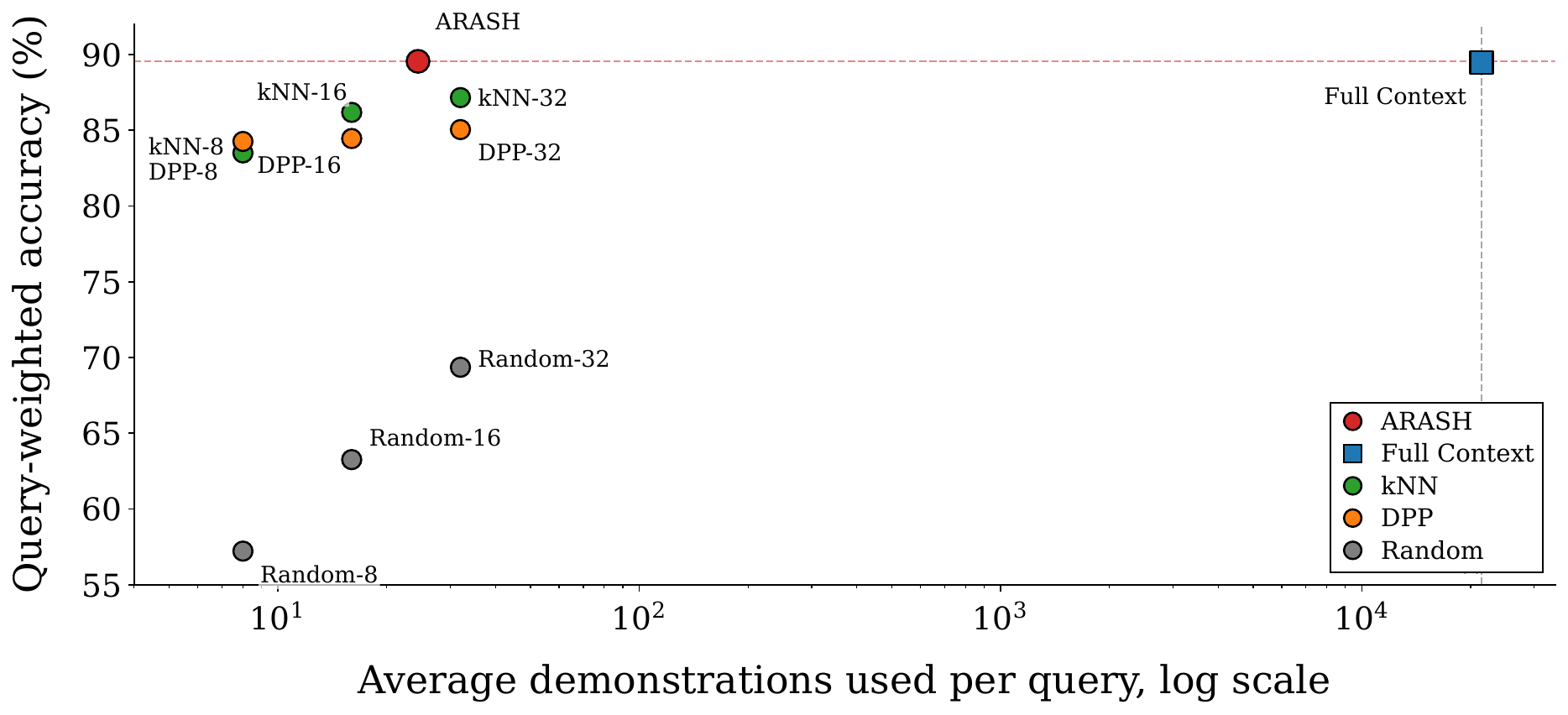}
        \caption{TabPFN}
        \label{fig:accuracy_vs_examples_tabpfn}
    \end{subfigure}
    %\hfill
    \begin{subfigure}[t]{0.48\textwidth}
        \centering
        \includegraphics[width=\linewidth]{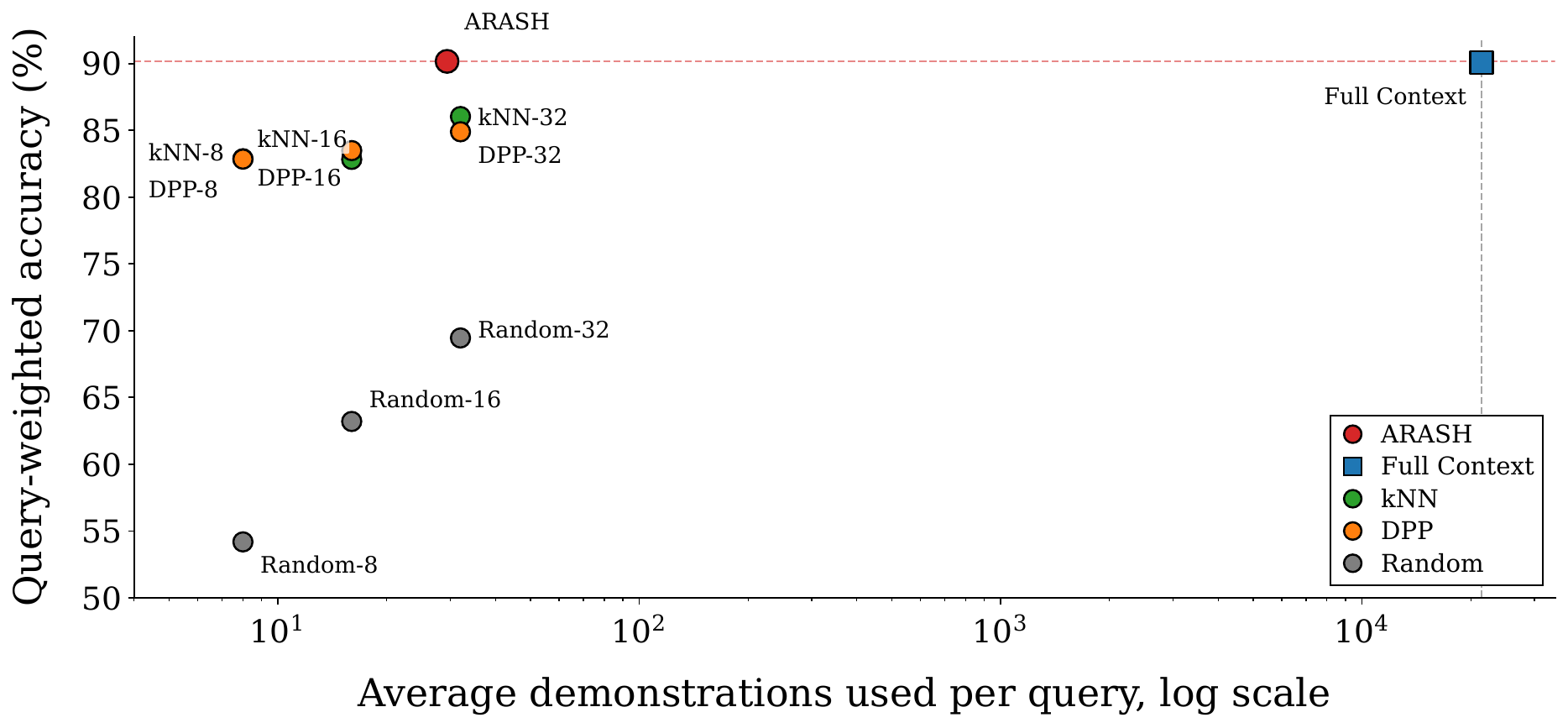}
        \caption{TabDPT}
        \label{fig:accuracy_vs_examples_tabdpt}
    \end{subfigure}

    \caption{Accuracy-efficiency comparison between \fname{} and ICL
    demonstration-selection baselines across (a)TabPFN and (b)TabDPT models. Lower number of demonstrations and higher accuracy are better. %The x-axis shows the average number of demonstrations used per query on a logarithmic scale, and the y-axis shows query-weighted accuracy. Full Context uses all available training
   % demonstrations, while kNN, DPP, and Random use fixed budgets. \fname{}achieves accuracy close to the full-context setting while using substantially fewer demonstrations.
    }
    \label{fig:accuracy_vs_examples}
\end{figure}

\begin{figure}[t]
    \centering
    \includegraphics[width=0.5\linewidth]{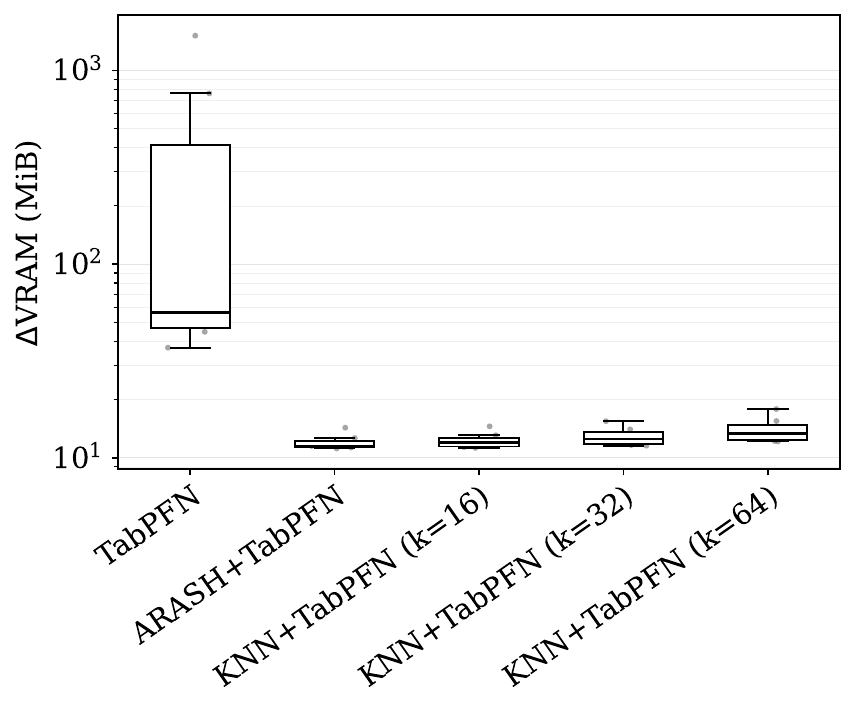}
    \caption{Inference-only peak $\Delta$VRAM in log scale measured during inference, aggregated across multiple datasets.
    %For each run, CUDA peak memory statistics are reset after \texttt{fit()}, and the maximum allocated memory during \texttt{predict()} is recorded and baseline-subtracted.
    %Methods include \textsc{TabPFN}, \textsc{ARASH+TabPFN}, and \textsc{KNN+TabPFN} with different values of $k$.
    %The y-axis is shown on a logarithmic scale.
    }
    \label{fig:vram}
\end{figure}

This section compares the accuracy and VRAM usage of ARASH with existing ICL techniques, kNN, DPP, random, and full context on two TFMs, TabPFN and TabDPT, shown in Figure~\ref{fig:accuracy_vs_examples}-\ref{fig:vram}. k in the three baselines, kNN-k, DPP-k, and Random-k, shows the selected demonstrations per query used in the baseline. The selected values of k=\{8, 16, 32\} are below and above the average number of demonstrations used by \fname{}, i.e., 24--30, for representative comparison.
\subsubsection{Accuracy}
 Both kNN and DPP improve accuracy as k increases, which suggests that local retrieval is useful for tabular ICL. However, their best fixed-budget results remain below \fname{}. For TabDPT, \fname{} achieves an accuracy of 0.9017 with 29.38 demonstrations per query on average, while kNN-32 and DPP-32 achieve an accuracy of 0.8603 and 0.8489, respectively. For TabPFN, \fname{}'s accuracy is 0.8955 with an average of 24.42 demonstrations per query, while kNN-32 and DPP-32 achieve an accuracy of 0.8716 and 0.8505, respectively. Random selection is consistently weaker, indicating that selecting relevant demonstrations to the query is as important as the number of demonstrations.
Full-context inference remains a high-cost baseline, achieving an average accuracy of 0.9009 for TabDPT and 0.8948 for TabPFN while using 30805 demonstrations per query. In contrast, \fname{} reaches a similar accuracy range with only a few dozen demonstrations per query, indicating the efficiency of ARASH.  %This suggests that \fname{} is not simply choosing a favorable fixed prompt size. Instead, it adapts both the number of demonstrations and the retrieval policy according to the locality and purity of each query region.
We also include a LoCalPFN-style large-budget retrieval setting for TabPFN as an additional reference point. Following the TabPFN-kNN inference rule used in LoCalPFN, we set $k_{\mathrm{LoCalPFN}} = \min(10\sqrt{N_{\mathrm{tr}}}, 1000)$
 where \(N_{\mathrm{tr}}\) is the number of training  demonstrations~\cite{thomas2024retrievalfinetuningincontext}. This rule makes the k dataset-dependent; larger training sets use a larger retrieval budget.
This makes the k value often large, thus some runs did not complete within our 20 min time limit. We therefore report this comparison only on the datasets for which both LoCalPFN-style kNN and DPP runs completed successfully. On these datasets, \(k_{\mathrm{LoCalPFN}}\) corresponds to 311.11 demonstrations per query on average. With this larger budget, LoCalPFN-style kNN and DPP obtain accuracies of 0.8821 and 0.8862, respectively. These results show that increasing k can improve fixed-budget retrieval, but the resulting accuracies remain below \fname{} in this comparison. In contrast, \fname{} uses a smaller adaptive k for each query with a higher accuracy.

%\Kazem{each paragraph should convey one message. You may highlight the message at the beginning of the paragraph}
\subsubsection{VRAM measurement}
To isolate memory usage induced specifically by the retrieved context at inference, we measure the \emph{predict-only peak} $\Delta$VRAM.
For each method, we first invoke \texttt{fit()} and record the GPU memory allocation immediately afterward.
We then reset CUDA peak-memory statistics and execute \texttt{predict()} on the test set, logging the maximum allocated memory during prediction.
$\Delta$VRAM is defined as the difference between this prediction-time peak and the post-fit baseline, thereby capturing only the additional allocations caused by the retrieved demonstrations.
This protocol explicitly excludes model parameters, one-time initialization overhead, and persistent allocator state shared across methods.

% \begin{figure}[t]
%   \centering
%   \includegraphics[width=0.3\linewidth]{figures/boxplot_delta_FLAN-T5.pdf}
%   \includegraphics[width=0.3\linewidth]{figures/boxplot_delta_LLaMA-3.2.pdf}
%   \includegraphics[width=0.3\linewidth]{figures/boxplot_delta_Qwen-2.5.pdf}
%   \caption{Accuracy improvement of \fname{} on LMs\Kazem{replot}%over the strongest baseline across datasets.
%   %Each boxplot shows the per-dataset difference in accuracy and Macro-F1
%   }
%   \label{fig:raf-vs-baseline}
% \end{figure}

\begin{figure}[t]
  \centering

  \begin{subfigure}{0.3\linewidth}
    \centering
    \includegraphics[width=\linewidth]{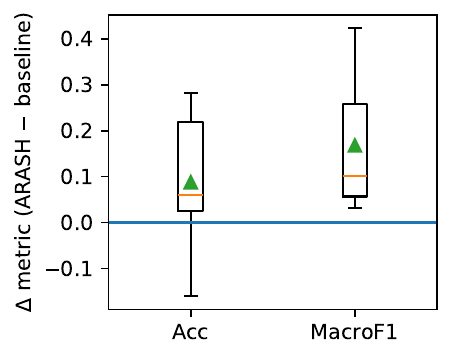}
    \caption{ LLaMA-3.2}
  \end{subfigure}
  \hfill
  \begin{subfigure}{0.3\linewidth}
    \centering
    \includegraphics[width=\linewidth]{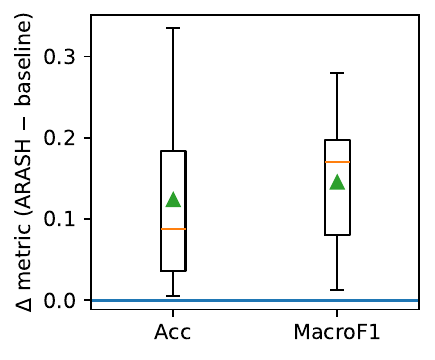}
    \caption{ Qwen-2.5}
  \end{subfigure}
  \hfill
  \begin{subfigure}{0.3\linewidth}
    \centering
    \includegraphics[width=\linewidth]{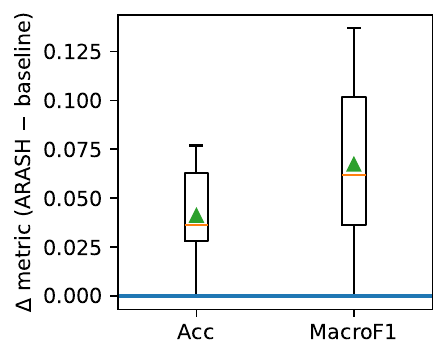}
    \caption{ FLAN-T5}
  \end{subfigure}

  \caption{Accuracy improvement of \fname{} over the strongest retrieval baseline across datasets for three language-model backbones.}
  \label{fig:raf-vs-baseline}
\end{figure}

% On the Combo dataset, Figure~\ref{fig:vram} shows that \textsc{ARASH} consistently incurs lower $\Delta$VRAM than \texttt{FULL\_TRAIN} across datasets, reflecting reduced memory pressure under adaptive retrieval.
% Compared to fixed-$k$ kNN+\textsc{TabPFN} baselines, \textsc{ARASH} typically matches the memory footprint of small-$k$ configurations while achieving higher predictive performance (Figure~\ref{fig:motiv}).
% As expected, the kNN baselines exhibit a monotonic increase in $\Delta$VRAM with $k$, consistent with linear growth in context length and activation memory.
% Overall, these results show that \textsc{ARASH} improves accuracy without increasing VRAM usage, allocating larger contexts only when they are required.
Figure~\ref{fig:vram} shows that \textsc{ARASH} consistently incurs lower $\Delta$VRAM than full context (TabPFN) across datasets, reflecting reduced memory pressure under adaptive retrieval.
As expected, the fixed-k baselines, only kNN is shown for brevity, exhibit a monotonic increase in $\Delta$VRAM with $k$, consistent with linear growth in context length and activation memory.
Overall, these results show that \textsc{ARASH} decreases VRAM usage, enabling running larger datasets on the same device.

\subsection{ARASH for Serialized-Table Language Models}
\label{sec:backends}

We further evaluate whether \fname{} generalizes beyond tabular foundation models by applying it to language-model backbones that operate on serialized tabular rows. As described in Section~\ref{sec:imp}, each tabular instance is converted into a textual representation, and demonstrations are provided to the model through an in-context prompt. We evaluate three LM backbones: FLAN-T5, LLaMA-3.2, and Qwen-2.5.
We compare \fname{} against three baselines that differ only in their demonstration-selection strategy, random selection, kNN across feature sets, and text. For each method, the number of demonstrations is constrained by the same prompt-token budget, so that the comparison reflects the quality of the selected demonstrations.

%Random selection samples demonstrations uniformly from the training set. %kNN-feature selects demonstrations using Euclidean distance in the standardized feature space. kNN-text selects demonstrations using cosine distance in the row-to-text embedding space. For each method, the number of demonstrations is constrained by the same prompt-token budget, so that the comparison reflects the quality of the selected demonstrations rather than a difference in context length.

Figure~\ref{fig:raf-vs-baseline} reports the distribution of performance gains achieved by \fname{} over the strongest retrieval baseline, shown separately for FLAN-T5, LLaMA-3.2, and Qwen-2.5. For each dataset, the baseline is defined as the best-performing method among random selection, kNN-feature retrieval, and kNN-text retrieval. The plotted values correspond to the per-dataset difference $\Delta = \text{\fname{}} - \text{baseline}$.
Across all three models, \fname{} achieves positive median gains over the strongest fixed retrieval baseline in both accuracy and Macro-F1. The median accuracy improvements are \(3.62\), \(5.99\), and \(8.77\) percentage points for
FLAN-T5, LLaMA-3.2, and Qwen-2.5, respectively. The corresponding Macro-F1 improvements are \(6.20\), \(10.07\), and \(17.00\) percentage points. These results show the portability of ARASH across LMs as well as TFMs. %that \fname{} provides systematic gains across LMs, with larger dispersion for LLaMA-3.2 and Qwen-2.5 than for FLAN-T5.
% $\Delta = \mathrm{Score}*{\fname{}} - \mathrm{Score}*{\mathrm{baseline}},

% where the score is accuracy or Macro-F1, depending on the metric being evaluated. Positive values indicate that \fname{} outperforms the strongest baseline on that dataset.

% Across all three backbones, \fname{} yields consistent positive improvements in accuracy and  Macro-F1. \Kazem{any number?}
% The median improvements are positive for all metrics, indicating that the gains are not driven by isolated datasets but reflect a systematic advantage over fixed retrieval strategies.
% Notably, gain dispersion varies across backbones. Although all three models show positive median improvements, differences in interquartile range and tail behavior are consistent with backbone-dependent variation in the distribution of performance gains across datasets.

\subsection{Analysis: When Does \fname{} Work?}
\label{sec:analysis-when-arash-helps}

% \fname{} reduces inference-time context by retrieving a compact, query-specific subset of training demonstrations rather than exposing the full-context set to the tabular foundation model. This reduction is beneficial when the retrieved context preserves the predictive evidence available in the Full-context set. We therefore analyze when cluster-local retrieval is reliable and when it is likely to discard useful information.

% This analysis follows the two assumptions used by \fname{} in Section~\ref{sec:arash}. First, the feature space should contain meaningful local neighborhoods, measured by the Hopkins locality score in Eq.~(2). Second, the routed local regions should provide consistent label evidence, measured by cluster purity and label entropy as defined in Step~II. Importantly, \fname{} does not require both diagnostics to be maximized. Even when locality is only moderate, ARASH can still be effective if the routed cluster has high purity, because the retrieved demonstrations provide a consistent class signal. In contrast, low-purity regions are more problematic because they indicate that local retrieval may mix conflicting labels and remove useful evidence from outside the routed cluster.

\begin{figure}[t]
    \centering
    \includegraphics[width=\linewidth]{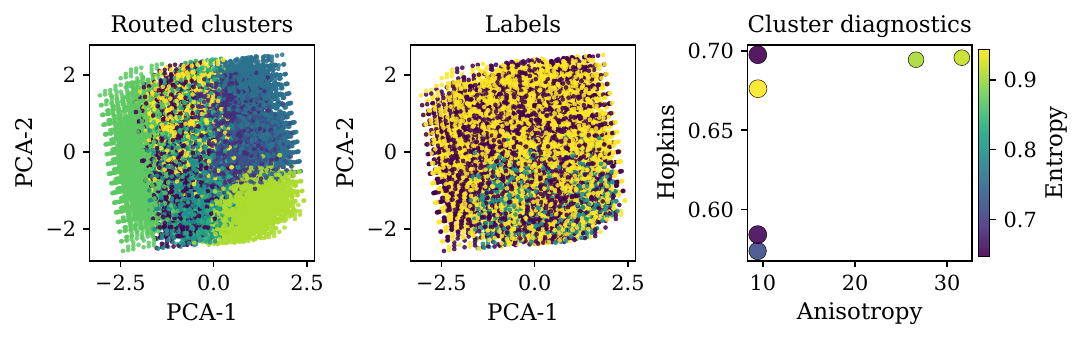}
    \vspace{0.5em}
    \includegraphics[width=\linewidth]{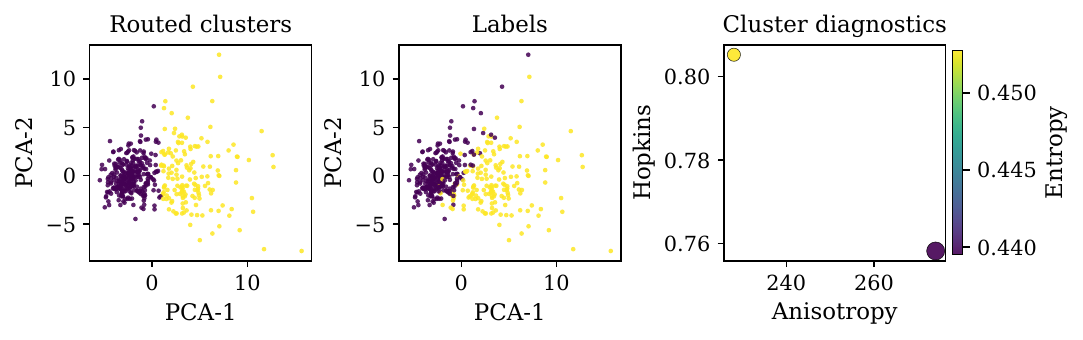}
    \caption{
    Representative analysis cases for ARASH. Top: Jungle. Bottom: WDBC. In each row, the left panel shows routed clusters in a two-dimensional PCA projection, the middle panel shows class labels in the same projection, and the right panel reports per-cluster diagnostics, including Hopkins locality, anisotropy, label entropy, and cluster size.
    % Representative diagnostic cases for ARASH.
    % Top: Churn. The routed clusters overlap in the PCA projection, and the label projection shows substantial class mixing, indicating weak support for cluster-local retrieval.
    % Bottom: Diabetes. The routed regions are more coherent, and the labels are more aligned with the local structure, indicating stronger support for compact local retrieval.
    % In each row, the left panel shows routed clusters in a 2D PCA projection, the middle panel shows labels in the same projection, and the right panel reports per-cluster diagnostics: Hopkins locality, anisotropy, label entropy, and cluster size.
    }
    \label{fig:jungle_diabetes_diagnostics}
\end{figure}

In Section~\ref{sec:motiv}, the key data patterns underlying the ARASH principles are demonstrated using synthetic data. These four synthetic regions are commonly found in real-world datasets. Specifically, across all queried datasets, we find that 43.2\%, 44.2\%, 9.7\%, and 2.9\% of queries belong to the high-locality/high-purity, high-locality/low-purity, low-locality/high-purity, and low-locality/low-purity groups, respectively. This distribution underscores the importance of the four distinct retrieval techniques. To further emphasize this point, this section extends the argument to two real-world datasets, Jungle and WDBC, as shown in Figure~\ref{fig:jungle_diabetes_diagnostics}, and then shows the applicability of the argument for other datasets.

\subsubsection{Representative cases.}

% \Samira{ I will update this part
% }
% These examples clarify the intended role of the diagnostics. \fname{} is not expected to improve every dataset merely by using fewer shots. It is expected to work when the query's routed region is informative enough to approximate the evidence provided by the Full-contex set. When locality is weak and labels are mixed within local regions, the selected context becomes less reliable. In such cases, Full-contex, global retrieval, or tree-based models may retain information that cluster-local retrieval discards.\Kazem{you put all this infor without providing any number! ok jungle is good but how much is its accuracy? what about others? kNN better? full context?}

%Figure~\ref{fig:jungle_diabetes_diagnostics} shows two representative datasets that illustrate the role of locality and label consistency in \fname{}. 
Jungle represents a difficult case for purely local retrieval. The routed clusters overlap substantially in the PCA projection, and the label projection shows strong class mixing within the same regions. The per-cluster diagnostics also show high entropy for some clusters. This indicates that locality alone is not sufficient: a cluster may contain nearby points, but if its label purity is weak, local nearest-neighbor retrieval can return demonstrations with conflicting labels.
This behavior motivates the retrieval decision in Step~III of \fname{}, which is a diversity-aware, hybrid, or global retrieval that may be discarded by a strictly local prompt.
%When the assigned cluster is not sufficiently pure, \fname{} should not rely only on locality. Instead, it can use diversity-aware, hybrid, or global retrieval to recover evidence that may be discarded by a strictly local prompt. 
The ARASH accuracy for Jungle supports this interpretation, where it matches the accuracy of full-context inference, which is \(0.965\). However, \fname{} uses only \(124\) demonstrations per query on average, compared with \(35{,}855\) training rows in the Full-context setting. In contrast, fixed-budget retrieval with \(k=128\) does not reach the same accuracy; the best DPP result reaches \(0.950\), and the corresponding kNN result remains below the Full-context and \fname{} accuracy. %Thus, Jungle shows that \fname{} can preserve Full-context-level performance with a much smaller prompt, while also illustrating why local kNN alone is not always reliable in low-purity regions.

% WDBC shows the favorable case for local retrieval. The routed regions exhibit clearer local structure, and the label projection is more consistent with the cluster organization. The per-cluster diagnostics also show lower label entropy, indicating that the local neighborhoods provide stable label evidence. In this regime, nearest-neighbor retrieval is less likely to select conflicting labels, so a compact retrieved prompt can preserve the relevant predictive information.The accuracy results are consistent with this diagnostic pattern. On WDBC, \fname{} matches Full-context inference, with both methods achieving (0.982) accuracy. Among the 167 test queries, \fname{} applies the direct-estimate path to 67 queries, with 100 accuracy on those direct predictions. Fixed-budget kNN retrieval also reaches (0.982) accuracy using (k=213) demonstrations, while the full training set contains 455 demonstrations.  This indicates that WDBC belongs to a favorable local-retrieval regime: the label signal is sufficiently concentrated that many queries can be resolved from a consistent local context, while the overall performance remains at the Full-context level.

The WDBC dataset represents a favorable regime for local retrieval, characterized by clear local structures, consistent label projections, and low per-cluster label entropy. These properties provide stable neighborhood evidence, reducing conflicting labels and allowing a compact retrieved prompt to preserve predictive information. 
Empirical results support this analysis. On WDBC, \fname{} matches full-context inference with 0.982 accuracy with only $k=12$ demonstrations. Out of 167 test queries, \fname{} routes 67 through the direct-estimate path, achieving 100\% accuracy on those predictions. Furthermore, fixed-budget kNN retrieval matches this 0.982 accuracy using only $k=213$ demonstrations out of the 455 available. This confirms that the dataset's label signal is sufficiently concentrated to allow accurate resolution from a localized context without sacrificing overall performance.

\subsubsection{Diagnostic association model.}

We next examine whether the qualitative behavior observed in the two case studies holds more broadly across datasets. 
\fname{} is based on the following mechanism: compact cluster-local prompts should be reliable when the feature space contains meaningful local neighborhoods and when the retrieved local regions provide consistent label evidence. If either condition is weak, local retrieval may discard useful global evidence or return conflicting demonstrations. We therefore study whether two mechanism-aligned diagnostics, dataset-level locality and final cluster purity, are associated with the accuracy gap between \fname{} and the Full-context setting.

For each dataset \(\mathcal{D}\) and foundation model \(\mathcal{M}\), where
\(\mathcal{M} \in \{\mathrm{TabPFN}, \mathrm{TabDPT}\}\), we define the accuracy gap as
% \begin{equation}
$\Delta_{\mathcal{D},\mathcal{M}}
=
\mathrm{Acc}^{\mathrm{ARASH}}_{\mathcal{D},\mathcal{M}}
-
\mathrm{Acc}^{\mathrm{Full}}_{\mathcal{D},\mathcal{M}}$
% \label{eq:arash_gap}
% \end{equation}
A negative value indicates that \fname{} is below Full-context accuracy, a value
close to zero indicates comparable accuracy, and a positive value indicates that
\fname{} exceeds Full-context accuracy while using fewer demonstrations.
We fit the following additive diagnostic model:
$\Delta_{\mathcal{D},\mathcal{M}}
=
\beta_0
+
\beta_L \mathrm{L}_{\mathcal{D}}
+
\beta_P P_\mathcal{D}
+
\epsilon_{\mathcal{D},\mathcal{M}}$
where \(\mathrm{L}_{\mathcal{D}}\) is the dataset-level locality score computed from the Hopkins
statistic in Eq.~(2), and \(P_\mathcal{D}\) is the average cluster purity for dataset \(\mathcal{D}\).
This model formalizes the same locality and purity conditions used by the
Algorithm\ref{alg:ss}. %It is used only as a diagnostic association model, not as a causalmodel. 
Since the observations are drawn from heterogeneous tabular datasets and
two foundation-model backbones, we report heteroskedasticity-robust standard
errors \citep{white1980heteroskedasticity}.
% Moreover, because some datasets
% contribute observations for more than one backbone, the estimates should be
% interpreted as descriptive evidence for the proposed mechanism rather than as
% independent causal effects.
%
Using the current TabPFN and TabDPT results, the fitted model is
% \begin{equation}
$\Delta_{\mathcal{D},\mathcal{M}}
=
-0.159
+
0.088 \mathrm{L}_{\mathcal{D}}
+
0.090 P_\mathcal{D} $.

% \label{eq:fitted_diagnostic_model}
% \end{equation}
Table~\ref{tab:diagnostic_regression} reports the coefficient estimates,
robust standard errors, and \(p\)-values over \(n=129\) observations. Both
coefficients are positive. After accounting for the other diagnostic, stronger
dataset-level locality and higher final cluster purity are each associated with
a smaller gap between \fname{} and the Full-context setting. This result supports
the intended operating regime of \fname{}: local prompting is most reliable when
local neighborhoods are meaningful and label-consistent.

\begin{table}[t]
\centering
\small
\setlength{\tabcolsep}{4pt}
\caption{
Diagnostic regression for the \fname{} accuracy gap
\(\Delta_{\mathcal{D},\mathcal{M}}=\mathrm{Acc}^{\mathrm{ARASH}}_{\mathcal{D},\mathcal{M}}
-\mathrm{Acc}^{\mathrm{Full}}_{\mathcal{D},\mathcal{M}}\).
Robust standard errors are reported over \(n=129\) observations.
}
\label{tab:diagnostic_regression}
\begin{tabular}{lccc}
\toprule
Predictor & Coef. & SE & \(p\) \\
\midrule
Dataset-level locality \(\mathrm{L}_{\mathcal{D}}\) & 0.088 & 0.027 & 0.0014 \\
Final purity \(P_\mathcal{D}\) & 0.090 & 0.019 & \(2.7{\times}10^{-6}\) \\
\bottomrule
\end{tabular}
\end{table}

The diagnostic association model provides empirical support for the Step~III thresholding strategy in Algorithm~\ref{alg:ss}. Step~III uses \(\tau_{\mathrm{loc}}\) and \(\tau_{\mathrm{pur}}\) to decide whether retrieval should remain compact and cluster-local or switch to diversity-aware, hybrid, or global retrieval. The positive coefficients for \(\mathrm{L}_{\mathcal{D}}\) and \(P_\mathcal{D}\) indicate that stronger locality and higher purity are associated with smaller \fname{}--Full-context accuracy gaps, which is consistent with the reliability assumptions used in Step~III.

\subsection{ARASH Implication on Inference Time}
%\subsection{Direct Local Prediction in Coherent Regions}
\label{sec:direct-local-prediction}
As discussed in Section~\ref{sec:cached_state}, the ARASH implementation uses a two-phase framework as well as a direct mode to reduce query execution time. This section discusses the impact of these techniques on overall prediction time. The preprocessing phase of ARASH is performed once its results are cached in a lookup table for reuse across subsequent queries. This design makes the overhead of table lookups negligible during inference, meaning that the total execution time is primarily dominated by the inference latency of the underlying model.

The inference cost of the underlying tabular foundation model increases with the number of demonstrations, although the exact scaling depends on the attention mechanism and model implementation details. \fname{} reduces this cost by using a smaller adaptive context. In the warm-query setting on the Combo dataset, full-context TabPFN uses an average of 600.25 demonstrations per query and requires 0.598 seconds per query, whereas \fname{} uses 13.69 demonstrations on average and requires 0.436 seconds per query, including amortized clustering, routing, retrieval, and local TFM inference. ARASH gives a 1.37$\times$ latency reduction while reducing the average context size by 43.9$\times$.
The absolute ARASH preprocessing time per query is less than 0.02 seconds. Most of the \fname{} query time is due to local TFM inference, which takes 0.422 seconds.   
%
%The non-model overhead is small: amortized clustering and preprocessing take 0.0119 seconds per query, routing takes 0.000011 seconds, retrieval takes 0.0011 seconds, and other overhead takes 0.00051 seconds. These results indicate that cached preprocessing and lookup-based routing introduce negligible per-query cost, and that inference latency remains dominated by the foundation-model call.

\fname{} further reduces inference cost through a direct path that bypasses model execution when the retrieved demonstrations are label-consistent. Across all evaluated queries, this direct path is used for 0.22 of the queries and achieves 0.989 accuracy. It is most frequent in the high-locality/high-purity regime, where it applies to 0.479 of the queries, compared with 0.156 when either diagnostic is low and 0.025 when both are low. This shows that coherent local structure can sometimes eliminate the need for a model inference, yielding near-zero model-inference cost for 22$\%$ of queries.

\subsection{Ablation Study}
\label{sec:ablation}
We ablate the main design choices in ARASH and particularly focus on two internal questions: whether the main components of ARASH are necessary, and whether both terms in the difficulty score contribute to shot allocation. We conduct the ablation study across the Combo dataset, while using the CREDIT dataset for detailed diagnostic illustration. CREDIT is selected because it provides a representative case with nontrivial local structure and mixed-label regions, and the same trend is observed across the other Combo datasets.

% \subsubsection{Adaptive shot allocation}
% ARASH assigns a query-specific number of demonstrations \(k_q\), rather than
% using the same number of shots for every test query. To isolate the effect of
% this design choice, we compare ARASH with fixed-\(k\) variants that use the
% same retrieval and ordering strategy but enforce a constant budget for all
% queries. ARASH is shown at its observed average budget,
% \(\bar{k}=14.03\), while the fixed baselines use
% \(k \in \{2,4,8,16,32,64\}\).

% Figure~\ref{fig:credit-adaptive-fixed} shows that ARASH achieves higher Macro-F1
% than the nearest fixed-budget baseline under a comparable average number of
% shots. This result indicates that the improvement is not explained only by
% increasing the context size. Instead, ARASH benefits from reallocating the
% available context across local regions: queries routed to more difficult
% regions receive more demonstrations, while queries in easier regions require
% fewer. This behavior is consistent with the objective of ARASH, which is to
% use the context budget selectively rather than uniformly.\Kazem{what is the message here? you had a more complete result before}

\subsubsection{Component ablation}
Table~\ref{tab:component_ablation} reports the accuracy of ARASH after removing one of its three steps.
Removing Step I, locality-aware clustering, reduces accuracy from \(0.750\) to
\(0.700\) and Macro-F1 from \(0.657\) to \(0.592\), showing that a global
retrieval policy is less effective than constructing prompts from local
regions. Replacing Step II, difficulty-aware allocation, with a fixed number of shots
per cluster also reduces the accuracy by 2\%, indicating that cluster-level difficulty
is useful for determining the amount of context. The largest degradation is
observed when informed retrieval is replaced with random retrieval, which
reduces Macro-F1 to \(0.495\).
These results support the design of ARASH as a coupled pipeline. Clustering
defines the candidate local regions, and difficulty-aware allocation determines
how much evidence each region requires, and retrieval selects demonstrations
that are relevant to the routed query. Removing any of these components
weakens the quality of the constructed prompt.

\begin{table}[t]
\centering
\caption{ The effect of the three steps of ARASH + TabpFN on the accuracy of the CREDIT dataset}
\label{tab:component_ablation}
\begin{tabular}{lcc}
\toprule
Method & Accuracy & Macro-F1 \\
\midrule
ARASH & 0.750 & 0.657 \\
ARASH Without Step I (No clustering) & 0.700 & 0.592 \\
ARASH Without Step II (Fixed shots per cluster) & 0.720 & 0.627 \\
ARASH Without Step III (Random retrieval) & 0.645 & 0.495 \\
\bottomrule
\end{tabular}
\end{table}

\subsubsection{Difficulty-score ablation}
Finally, we examine the two terms used in the cluster difficulty score.
For a cluster \(c\), ARASH computes
$d_c = \alpha H_c + (1-\alpha) I_c$
where \(H_c\) is the normalized label entropy and
\(I_c = 1-P_c\) is the cluster impurity. Entropy measures uncertainty over
the full label distribution, whereas impurity measures the absence of a
dominant local label. These two quantities are related but not identical:
Two clusters may have the same majority-class fraction while differing in how
the remaining labels are distributed.
Table~\ref{tab:difficulty_ablation} compares entropy-only, impurity-only,
and combined difficulty scores under identical train--test splits across five
seeds. Using either term alone reduces both Accuracy and Macro-F1. The
combined score gives the best result, suggesting that entropy and impurity
provide complementary information for adaptive shot allocation. This supports
the use of both terms in ARASH: entropy captures distributional uncertainty,
while impurity captures local label conflict relative to the dominant class.

\begin{table}[t]
\centering
\caption{The effect of difficulty-score component on ARASH + TabPFN accuracy for the CREDIT dataset}
\label{tab:difficulty_ablation}
\begin{tabular}{lcc}
\toprule
Variant & Accuracy & Macro-F1 \\
\midrule
ARASH with Entropy only score & \(0.710 \pm 0.024\) & \(0.630 \pm 0.038\) \\
ARASH with Impurity only score & \(0.720 \pm 0.020\) & \(0.640 \pm 0.026\) \\
ARASH with Entropy + impurity & \(0.753 \pm 0.010\) & \(0.671 \pm 0.029\) \\
\bottomrule
\end{tabular}
\end{table}

% \begin{figure}[t]
%   \centering
%   \includegraphics[width=\columnwidth]{figs/vldb_credit_macroF1.pdf}
%   \caption{
%   Adaptive per-query shot allocation versus fixed-$k$ retrieval on \textsc{Credit}. 
%   Fixed-$k$ baselines use the same number of retrieved examples for every query. 
%   ARASH selects a query-specific budget $k_q$ and is shown at its average budget $\bar{k}=14.03$.
%   }
%   \label{fig:credit-adaptive-fixed}
% \end{figure}

\section{Related Work}

Classical tree-based and neural methods train models for a given tabular data. 
Tabular prediction is dominated by tree ensembles, especially gradient-boosted decision trees, such as XGBoost~\cite{chen2016xgboost}. These models handle heterogeneous features, missing values, small data, and non-smooth boundaries well.

Neural models for tabular data include attention-based architectures such as TabTransformer~\cite{huang2020tabtransformer}, FT-Transformer~\cite{gorishniy2021revisiting}, SAINT~\cite{somepalli2021saint}, and AutoInt~\cite{song2019autoint}, which model feature interactions, and methods like  TabNet~\cite{arik2021tabnet}, DANets~\cite{chen2021danets}, TANGOS~\cite{jeffares2023tangos}, and SwitchTab~\cite{wu2024switchtab}, which introduce structured biases or self-supervision. These approaches train model parameters for a target dataset. In contrast, ARASH studies inference with a frozen model, focusing on selecting labeled rows, their regions, and query-specific shot budgets.

% Large language models introduced ICL, where task adaptation arises from demonstrations in the prompt rather than parameter updates~\cite{brown2020language}. Prior work shows ICL performance is sensitive to demonstration choice~\cite{liu2022makes}, motivating retrieval-based ICL that selects query-relevant examples instead of fixed or random prompts, including learned retrievers~\cite{rubin2022learning}, nearest-neighbor prompting~\cite{shi2022knn,xu2023k}, and broader retrieval-augmented strategies~\cite{luo2024retrievedsurvey}. Diversity-aware methods further improve selection: determinantal point processes provide a principled way to select diverse subsets~\cite{kulesza2012determinantal} and have been applied to representative in-context examples~\cite{yang2023representative}, while clustering-based approaches select demonstrations that are both similar to the query and diverse~\cite{li2025quickly}. These techniques are primarily designed for non-tabular data with natural geometric locality, whereas tabular data has arbitrary row order, making locality and purity harder to define. 

% Large language models introduced ICL, where task adaptation arises from demonstrations in the prompt rather than parameter updates~\cite{brown2020language}. Prior work shows ICL performance is sensitive to demonstration choice~\cite{liu2022makes}, motivating retrieval-based ICL that selects query-relevant examples instead of fixed or random prompts,including  nearest-neighbor prompting~\cite{shi2022knn,xu2023k}.

Large language models introduced ICL, in which task adaptation is performed through demonstrations provided in the prompt rather than through parameter updates~\cite{brown2020language}. Prior work has shown that ICL performance is sensitive to the choice of demonstrations~\cite{liu2022makes}. This observation has motivated retrieval-based ICL methods, which select query-relevant examples instead of relying on fixed or random prompts, including nearest-neighbor prompting~\cite{shi2022knn,xu2023k}.
Diversity-aware methods further improve selection: determinantal point processes provide a principled way to select diverse subsets~\cite{kulesza2012determinantal} and have been applied to representative in-context examples~\cite{yang2023representative}, while clustering-based approaches select demonstrations that are both similar to the query and diverse~\cite{li2025quickly}. These techniques are primarily designed for non-tabular data with natural geometric locality, whereas tabular data has arbitrary row order, making locality and purity harder to define. 

%\subsection{Tabular language models and tabular foundation models}
Recent work extends foundation-model approaches to tabular prediction along two directions. The first serializes tabular rows into text and applies LMs to prompts. TabLLM~\cite{hegselmann2023tabllm} converts rows into textual descriptions for classification; this is flexible due to the language interface, but becomes expensive when many rows are included. The second develops table-native foundation models. TabPFN~\cite{hollmann2022tabpfn} formulates classification as prior-data-fitted inference using labeled context examples, TabDPT~\cite{ma2024tabdpt} studies scalable tabular foundation modeling, and TabICL~\cite{qu2025tabicl} scales tabular ICL via fixed-dimensional embeddings followed by transformer-based prediction. These approaches directly target tabular prediction and cross-table transfer. While effective, building a new model is expensive and needs huge computational resources. 

A closely related line studies retrieval and context construction for tabular prompting. Black-box prompting techniques such as full-context methods~\cite{hollmann2022tabpfn} include all demonstrations regardless of the query or training set. While simple, this approach incurs unnecessary computation and memory costs and can introduce irrelevant information into the prompt, potentially degrading performance. Another category constructs prompts conditioned on the query. These methods use a fixed budget~\cite{wu2025efficient} and select demonstrations either randomly or via fixed-$k$ nearest-neighbor prompting~\cite{shi2022knn,xu2023k,wu2025efficient}. Although they reduce prompt length by focusing on local demonstrations, their accuracy may degrade when local regions exhibit impure label distributions. \fname{} constructs prompts based on both locality and purity, enabling more effective context selection tailored to tabular data.

\section{Conclusion}
This work introduces ARASH, an adaptive retrieval and shot selection method for tabular prediction tasks using TFMs and LMs. ARASH extracts statistical features from datasets to establish locality and purity for demonstrations, subsequently augmenting the context provided to the query. This is achieved through a three-stage process: locality-aware clustering, difficulty-aware shot selection, and a retrieval technique. %These steps are implemented as pre-processing and inference stages to improve the efficiency of ARASH.
We evaluated ARASH using TabPFN, TabDPT, and several LMs; experimental results demonstrate that ARASH achieves comparable accuracy to full context while significantly reducing token length. Consequently, prompt length and VRAM usage are reduced by 1261.5$\times$ and 2.56$\times$ across the datasets.

\bibliographystyle{IEEEtran} %default: IEEEtran
\bibliography{new_refs}

 %\appendix

\end{document}